\documentclass[preprint,12pt]{elsarticle}

\usepackage{amssymb}
\usepackage{amsmath}

\usepackage[utf8]{inputenc} % allow utf-8 input
\usepackage[T1]{fontenc}    % use 8-bit T1 fonts
\usepackage{hyperref}       % hyperlinks
\usepackage{url}            % simple URL typesetting
\usepackage{booktabs}       % professional-quality tables
\usepackage{amsfonts}       % blackboard math symbols
\usepackage{nicefrac}       % compact symbols for 1/2, etc.
\usepackage{lipsum}
\usepackage{graphicx}
\usepackage{amsmath}

\usepackage{footmisc}

\usepackage{makecell}

\usepackage[utf8]{inputenc} % allow utf-8 input
\usepackage[T1]{fontenc}    % use 8-bit T1 fonts
\usepackage{url}            % simple URL typesetting
\usepackage{booktabs}       % professional-quality tables
\usepackage{amsfonts}       % blackboard math symbols
\usepackage{nicefrac}       % compact symbols for 1/2, etc.
\usepackage{xcolor}         % colors
\usepackage{tabularx}
\usepackage{subcaption}
\usepackage{multicol}
\usepackage{multirow}
\usepackage{tabularray}
\usepackage{amsmath}
\usepackage{amssymb}
\usepackage{siunitx}
\usepackage{colortbl}
\usepackage{graphicx}
\usepackage{lipsum}
\usepackage{hyperref}       % hyperlinks
\usepackage[capitalize]{cleveref}
\usepackage[export]{adjustbox}
\usepackage{wrapfig}
\usepackage{svg}
\usepackage{amssymb}
\usepackage{sidecap}
\usepackage{caption}
\usepackage{pifont}

\journal{Information Fusion}
\begin{document}



\title{HIDFlowNet: A Flow-Based Deep Network for Hyperspectral Image Denoising} %% Article title

%%\author[label1,label2]{}
\author[1]{Qizhou Wang\fnref{2}}
\ead{qzwang@stu.xjtu.edu.cn}

\author[1]{Li Pang\fnref{2}}
\ead{2195112306@stu.xjtu.edu.cn}

\author[1]{Xiangyong Cao\corref{cor1}}
\ead{caoxiangyong@xjtu.edu.cn}

%\author[1]{Weizhen Gu}
%\ead{1911693@mail.nankai.edu.cn}

%\author[1]{Xiangyu Rui}
%\ead{xyrui.aca@gmail.com}

%\author[1]{Jiangjun Peng}
%\ead{andrew.pengjj@gmail.com}

%\author[1]{Shuang Xu}
%\ead{andrew.pengjj@gmail.com}

\author[1]{Zhiqiang Tian\corref{cor1}}
\ead{zhiqiangtian@xjtu.edu.cn}

\author[1]{Deyu Meng}
\ead{dymeng@xjtu.edu.cn}

\cortext[cor1]{Corresponding author. $^1$Co-first Author.}

\affiliation[1]{organization={Xi'an Jiaotong University},
            addressline={Xianning West Road},
            city={Xi'an},
            postcode={710049},
            state={Shaanxi},
            country={China}}

%\author{Jin Cao, Xiangyu Rui, Li Pang, Xiangyong Cao} %% Author name

%\affiliation{organization={csac},%Department and Organization
   %         addressline={},
    %        city={},
     %       postcode={},
      %      state={},
       %     country={}}

%% Abstract
\begin{abstract}
Hyperspectral image (HSI) denoising is essentially ill-posed since a noisy HSI can be degraded from multiple clean HSIs. However, existing deep learning (DL)-based approaches only restore one clean HSI from the given noisy HSI with a deterministic mapping, thus ignoring the ill-posed issue and always resulting in an over-smoothing problem. Additionally, these DL-based methods often neglect that noise is part of the high-frequency component and their network architectures fail to decouple the learning of low-frequency and high-frequency. To alleviate these issues, this paper proposes a flow-based HSI denoising network (HIDFlowNet) to directly learn the conditional distribution of the clean HSI given the noisy HSI and thus diverse clean HSIs can be sampled from the conditional distribution. Overall, our HIDFlowNet is induced from the generative flow model and is comprised of an invertible decoder and a conditional encoder, which can explicitly decouple the learning of low-frequency and high-frequency information of HSI. Specifically, the invertible decoder is built by staking a succession of invertible conditional blocks (ICBs) to capture the local high-frequency details. The conditional encoder utilizes down-sampling operations to obtain low-resolution images and uses transformers to capture correlations over a long distance so that global low-frequency information can be effectively extracted. Extensive experiments on simulated and real HSI datasets verify that our proposed HIDFlowNet can obtain better or comparable results compared with other state-of-the-art methods.
\end{abstract}

%% Please copy and paste the code instead of the example below
\maketitle

\section{Introduction}
Hyperspectral image (HSI) depicts an object in numerous narrow and contiguous spectral bands across the electromagnetic spectrum. Compared with RGB images, HSIs enable a more comprehensive depiction of captured scenes due to hundreds of spectral bands and have been widely applied in various applications, such as classification~\cite{cao2018hyperspectral,cao2020hyperspectral}, object detection~\cite{liu2021sraf}, medical diagnosis~\cite{calin2014hyperspectral}, agriculture~\cite{dale2013hyperspectral} and so on. However, owing to multiple factors such as instrument instability, circuit malfunction and light disturbance, HSIs are often subjected to various noises during the data acquisition stage, which can negatively impact the performance of the downstream applications. Therefore, the HSI denoising task is an important pre-processing step in HSI analysis. Recently, numerous HSI denoising techniques have been proposed and these methods can be categorized into two classes, i.e., model-based approaches and deep learning-based methods.

\begin{figure*}
  \centering
  \includegraphics[width=\textwidth]{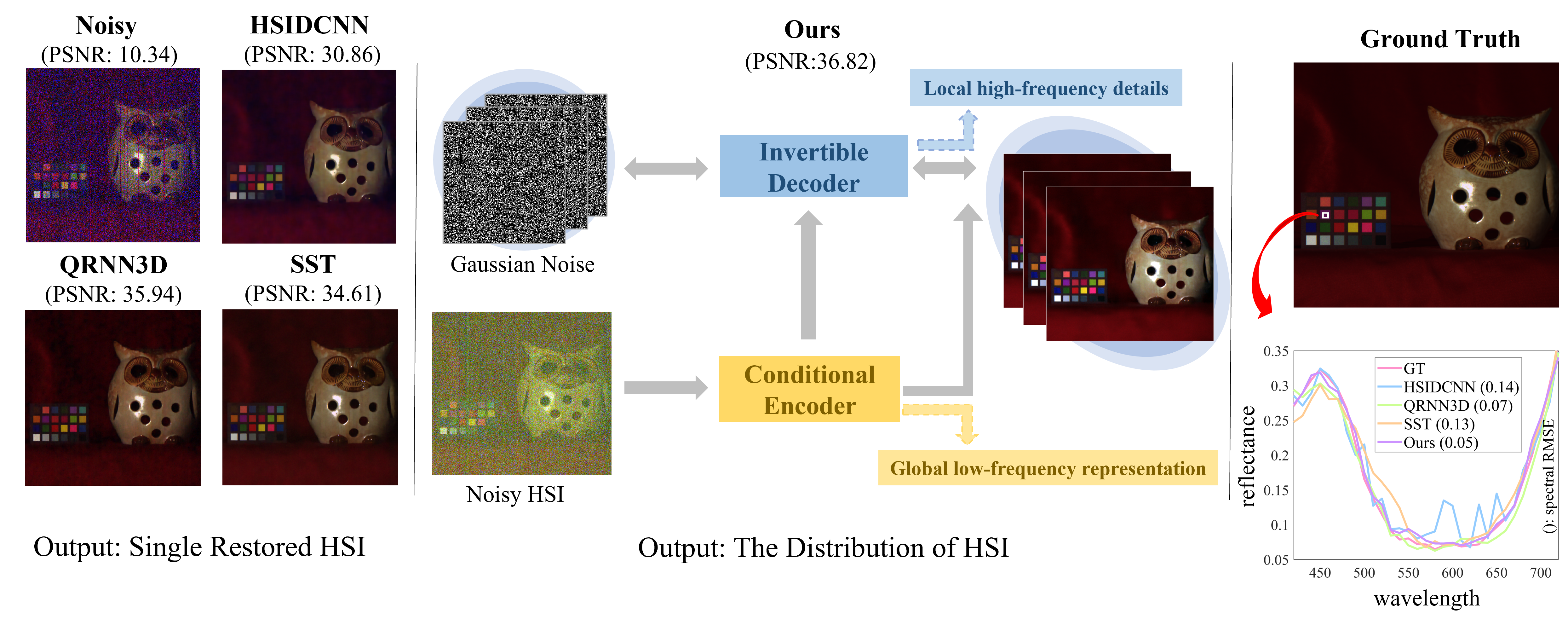}
  \caption{Instead of performing HSI denoising with a deterministic mapping, our HIDFlowNet learns the conditional distribution of clean HSI given corresponding noisy counterpart, which explicitly alleviates the ill-posed nature of HSI denoising and enables us to sample diverse clean HSIs. The charts on the right demonstrate that the reconstructed spectral reflectance of our HIDFlowNet is more consistent with the ground truth than that of other approaches (the spectral RMSE of our method is 0.05 while that of the second-best method QRNN3D is 0.07).}
  \label{fig:teaser}
\end{figure*}

Model-based HSI denoising approaches focus on exploiting the prior knowledge of HSIs, such as spatial-spectral total variation~\cite{He2016Total}, spatial-spectral non-local mean~\cite{Qian2013Hyperspectral}, spatial-spectral sparse representation~\cite{Peng2014Decomposable}, and low-rank prior~\cite{chen2017denoising,chen2018denoising}, and are implemented in an iterative optimization manner. However, since the characteristics of HSIs are complex, the hand-crafted priors thus can only partially reflect the property of HSIs, making these approaches incapable of handling unknown real HSI. Moreover, the iterative optimization process of denoising a single HSI consumes a significant amount of time. In contrast, by utilizing the impressive non-linearity capability of neural networks, deep learning (DL)-based approaches are capable of capturing the intrinsic characteristics of HSIs in a data-driven manner and can also learn the underlying image features statistically with abundant clean and noisy image pairs. Although these approaches can achieve desirable performance, they can only predict one clean HSI for a given noisy HSI with a deterministic mapping (see Fig.~\ref{fig:teaser}), thus ignoring the ill-posed nature of HSI denoising task, i.e., a noisy HSI can be degraded from multiple clean HSIs. Besides, these deterministic DL-based methods overemphasize pixel similarity and tend to predict the average of all possible clean images, resulting in over-smoothed areas and loss of image details~\cite{lugmayr2020srflow}. Although adversarial training~\cite{zhang2022hyperspectral} and perceptual loss~\cite{gao2021hyperspectral} have been adopted to alleviate this problem, these methods still do not address the ill-posed issue as they only predict a single clean HSI. Additionally, most existing DL-based methods~\cite{chang2018hsi,yuan2018hyperspectral,wei20203,pan2022sqad} often neglect the fact that noise is part of the high-frequency component and thus their network architectures fail to decouple the learning of low-frequency and high-frequency, which make these networks lack specific physical interpretations.

To alleviate these issues, this paper proposes a flow-based hyperspectral image denoising network (i.e., HIDFlowNet), which aims to directly learn the conditional distribution of the clean HSIs by transforming the unknown conditional distribution of clean HSIs given the noisy HSIs into a known Gaussian distribution (see Fig.~\ref{fig:teaser}). Moreover, the
proposed HIDFlowNet is capable of decoupling the learning of low-frequency and high-frequency information of HSI by its two main components: a conditional encoder network and an invertible decoder network. The encoder network composed of a series of transformer blocks and down-sampling operations, is utilized to extract global low-frequency information in an unsupervised manner. More specifically, the down-sampling operations employed in the encoder enable the network to obtain low-resolution images so that low-frequency details can be extracted efficiently. Also, transformers which are able to capture long-distance correlations are adopted to extract global information effectively. Moreover, the invertible decoder is built by staking a successive of invertible conditional blocks (ICBs) to preserve local high-frequency details since invertible networks are information-lossless~\cite{liu2020deep}. Finally, HIDFlowNet is trained by minimizing the negative log-likelihood of the conditional distribution and a reconstruction loss to obtain high-quality HSIs. Once the training is finished, diverse clean HSIs corresponding to one noisy HSI can be generated by first sampling in the latent space and then performing inverse transforms. Sampling multiple times can ensure the diversity of generated clean images as different sampled images are probable to focus on different detailed parts of the ground-truth clean HSI.

In summary, our contributions are shown as follows:
\begin{itemize}
\item A flow-based network namely HIDFlowNet, which learns a bijective mapping between a simple Gaussian distribution and a complex data distribution, is proposed to learn the conditional distribution of a clean HSI given its corresponding noisy counterpart. The model is able to generate diverse restored images by sampling random Gaussian noise and performing inverse transforms. To our knowledge, this is the first attempt to employ a flow-based model for HSI denoising.
\item The architecture of HIDFlowNet induced from the flow methodology contains two main components and has an explicit physical interpretation since it decouples the learning of low-frequency and high-frequency information of HSI. The invertible decoder preserves the local high-frequency details and the conditional encoder network extracts global low-frequency representation. Two main components enables the network to enhance low-frequency and high-frequency information simultaneously. 
\item Extensive experiments on the simulated and real HSI datasets verify the superiority of our proposed method compared with other state-of-the-art methods.
\end{itemize}

\section{Related Works}
In this section, we give a brief review of several research fields related to our work, including HSI denoising approaches and flow-based generative models.

%\emph{Model-based Methods.}
\subsection{Model-based HSI Denoising Methods}
Model-based HSI denoising methods utilize priori information about the underlying statistical properties of the hyperspectral data to perform denoising. Handcrafted priors such as low-rank \cite{li2015hyperspectral,chang2017hyper, fan2017hyperspectral,cao2016robust,9975834,chen2017denoising,peng2022fast}, sparse representation \cite{xie2016multispectral, xue2021spatial, zhao2014hyperspectral, ma2019local}, total variation \cite{yuan2012hyperspectral, he2015total, he2018hyperspectral} and nonlocal similarity \cite{he2019non} are proposed and corresponding model regularization terms are designed to obtain promising denoising results. For example, in \cite{zhang2013hyperspectral}, low-rank matrix recovery (LRMR) is proposed to simultaneously remove various noises by utilizing the low-rank property of HSIs and the sparsity nature of non-Gaussian noise. Cao \emph{et al.} \cite{cao2016robust} proposed a mixture of exponential power distribution in the low-rank matrix factorization framework to capture the complex noise of HSIs. Xue \emph{et al.} \cite{xue2021spatial} proposed a structured sparse low-rank representation (SSLRR) model to induce sparse property. Spatial-spectral total variation regularized local low-rank matrix recovery (LLRSSTV) \cite{he2018hyperspectral} employed a global reconstruction strategy to fully utilize both low-rank property and smoothness properties of HSIs. He \emph{et al.} \cite{he2019non} proposed NGMeet which unified spatial and spectral low-rank properties. Although these methods can effectively preserve the spectral and spatial characteristics of HSIs, the optimization of the model is very complex and thus these methods are always time-consuming. In addition, the denoising performance is highly dependent on the consistency between the priors and HSIs. However, manually designed priors only reflect the intrinsic characteristics of HSIs partially, limiting their ability for HSI denoising.

\subsection{Deep Learning-based HSI Denoising Methods}
Recently, deep learning-based methods for HSI denoising gain increasing attention and popularity owing to the powerful nonlinear fitting ability of neural networks. These methods capture the statistical characteristics of HSIs in a data-driven manner with a large number of training pairs. For instance, HSI-DeNet~\cite{chang2018hsi} employs a 2-D convolutional neural network to learn multiple image filters for HSI denoising. HSID-CNN \cite{yuan2018hyperspectral} employs convolution kernels of multiple sizes to extract multilevel features, which are then fused to restore the HSIs. QRNN3D \cite{wei20203} introduces 3-D convolution blocks and quasi-recurrent mechanisms to extract spatial and spectral simultaneously without damaging the image structure. SQAD~\cite{pan2022sqad} designed a spatial-spectral quasi-recurrent attention unit (QARU) to maintain high-quality spatial and spectral information. GRN~\cite{cao2022deep} used two reasoning modules based on the graph neural network (GNN) to carefully extract both global and local spatial-spectral features. TRQ3DNet~ \cite{pang2022trq3dnet} first introduces a vision Transformer in HSI denoising, modelling the spatial long-range dependencies of HSIs and achieving desirable denoising performance. SST~\cite{li2022spatial} conducts attention mechanisms in both spatial and spectral dimensions to fully explore the similarity characteristics of HSIs. HWnet~\cite{rui2022hyper} is proposed to improve the generalization ability of model-based methods in a data-driven manner. While demonstrating promising denoising performance, these approaches learn a deterministic mapping and neglect the fundamental ill-posed nature of HSI denoising.

\subsection{Flow-based Generative Models}
Flow-based generative models have shown promising results in a variety of applications, including image generation \cite{ren2020deep}, speech synthesis \cite{cong2021glow}, and physics simulations \cite{deng2020modeling}. These models transform a complex distribution into a known simple distribution (e.g., Gaussian Distribution) with an invertible network so that diverse samples can be obtained by sampling in the known latent space and performing inverse transforms. For example, NICE \cite{dinh2014nice} stacks several additive coupling layers and a rescaling layer to learn manifolds. Based on NICE, RealNVP \cite{dinh2016density} further proposes affine coupling layers with masked convolution to improve fitting ability. Glow \cite{kingma2018glow} employs invertible 1$\times$1 convolutions to perform channel permutations and actnorm layers to accelerate training. Recently, flow-based models have been increasingly applied to various computer vision tasks. For example, SRFlow \cite{lugmayr2020srflow} models the conditional distribution of high-resolution images given corresponding low-resolution images, enabling the trained model to predict diverse high-resolution photos. VideoFlow \cite{kumar2019videoflow} predicts high-quality stochastic multi-frame videos based on past observations using a normalizing flow. In this paper, we follow this research line and further exploit the application of flow-based methods in the HSI denoising task.

\section{The Proposed Method}
In this section, we provide a detailed description of our proposed HIDFlowNet. Firstly, we present the problem of the ill-posed nature of HSI denoising and then introduce conditional flow models. Next, we illustrate the network structure of HIDFlowNet in detail.

\subsection{Conditional Generative Flows}
\label{sec:flow}
The task of HSI denoising is to restore clean HSIs from given noisy HSIs. Generally, a degraded HSI can be mathematically modeled as
\begin{equation}\label{eq:NoisyHSI}
\mathbf{Y} = \mathbf{X} + \epsilon,
\end{equation}
where $\mathbf{Y} \in \mathbb{R}^{H\times W\times B}$ denotes the degraded HSI, $\mathbf{X} \in \mathbb{R}^{H\times W\times B}$ is the corresponding clean HSI and $\epsilon \in \mathbb{R}^{H\times W\times B}$ stands for the additive noise. $H, W$ and $B$ denote the height, width and spectral band number of the HSI, respectively.

\begin{figure*}[h]
  \centering
  \includegraphics[width=0.95\linewidth]{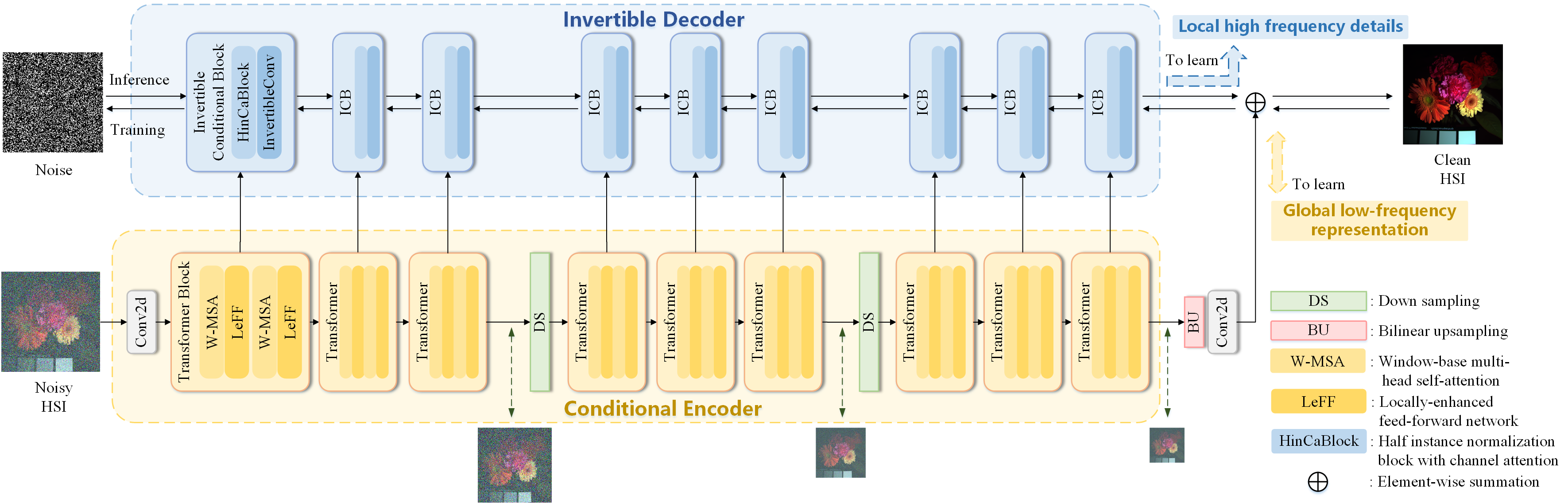}
  \caption{The network architecture of HIDFlowNet includes a conditional encoder (yellow) and an invertible decoder (blue). The encoder takes the noisy HSI as input and generates multiple-scale feature maps with a series of transformer blocks and down-sampling operations. The invertible decoder transforms a latent representation which conforms to a simple distribution (e.g., a Gaussian distribution) into high-frequency information utilizing a succession of invertible conditional blocks with the guidance of the encoder. Finally, the low and high-frequency parts are merged to restore clean HSI. The whole framework is trained by minimizing the negative log-likelihood and reconstruction loss, and then can predict diverse clean HSIs during the inference stage.}
  \label{fig:architecture}
\end{figure*}

As previously mentioned, HSI denoising is an ill-posed problem since a noisy HSI can be degraded from multiple clean HSIs that are equally reasonable. Therefore, instead of learning a deterministic mapping $\mathbf{Y} \rightarrow \mathbf{X}$ as existing deep learning-based methods do, we propose to employ a flow-based network $f_{\theta}$ to learn the conditional distribution $P_{\mathbf{X}|\mathbf{Y}}(\mathbf{X}|\mathbf{Y}, \boldsymbol{\theta})$ of clean HSI $\mathbf{X}$ given corresponding noisy counterpart $\mathbf{Y}$. Specifically, the network is designed to be invertible to guarantee one-to-one mapping. To put it another way, the invertible network transforms a clean and noisy HSI pair $(\mathbf{X}, \mathbf{Y})$ into a latent variable $\mathbf{z} = f_{\boldsymbol{\theta}}(\mathbf{X};\mathbf{Y})$, and the clean HSI $\mathbf{X}$ can be reconstructed exactly by performing inverse transforms as $\mathbf{X} = f^{-1}_{\boldsymbol{\theta}}(\mathbf{z};\mathbf{Y})$. In this context, by applying the change-of-variables formula, the probability density of $p_{\mathbf{X}|\mathbf{Y}}$ can be explicitly defined as
\begin{equation}\label{eq:pd}
p_{\mathbf{X}|\mathbf{Y}}(\mathbf{X}|\mathbf{Y},\boldsymbol{\theta})=p_{\mathbf{z}}\big(f_{\boldsymbol{\theta}}(\mathbf{X};\mathbf{Y})\big)\left|\det\dfrac{\partial f_{\boldsymbol{\theta}}}{\partial\mathbf{X}}(\mathbf{X};\mathbf{Y})\right|,
\end{equation}
where the $det(\cdot)$ term is the determinant of the Jacobian matrix $\dfrac{\partial f_{\boldsymbol{\theta}}}{\partial\mathbf{X}}(\mathbf{X};\mathbf{Y})$.
Therefore, the conditional distribution of the clean HSI can be directly learned by minimizing the negative log-likelihood (NLL) as
\begin{equation}
\label{eq:nll}
\begin{split}
\mathcal{L}_{nll}(\boldsymbol{\theta};\mathbf{X},\mathbf{Y}) =& -\log p_{\mathbf{X}|\mathbf{Y}}(\mathbf{X}|\mathbf{Y},\boldsymbol{\theta}) \\
=& -\log p_\mathbf{z}\big(f_{\boldsymbol{\theta}}(\mathbf{X};\mathbf{Y})\big)-\log\left|\det\dfrac{\partial f_{\boldsymbol{\theta}}}{\partial\mathbf{X}}(\mathbf{X};\mathbf{Y})\right|.
\end{split}
\end{equation}
Moreover, the flow-based network is decomposed into a succession of invertible layers so that the determinant term in Eq.(\ref{eq:nll}) can be readily calculated. Specifically, the flow-based network consists of $N$ invertible layers, $\mathop{\mathrm{i.e.,}}f_{\boldsymbol{\theta}}=f_{\boldsymbol{\theta}}^{N}f_{\boldsymbol{\theta}}^{N-1}\cdots f_{\boldsymbol{\theta}}^{1}$, where $f_{\boldsymbol{\theta}}^{n}$ denotes the $n_{th}$ layer. The $n_{th}$ layer takes the outputs of the previous layer as inputs, $\mathop{\mathrm{i.e.,}} \mathbf{h}^{n+1} = f^{n}_\theta(\mathbf{h}^n;\mathbf{X})$, where $\mathbf{h}^{1} = \mathbf{X}$ and $\mathbf{h}^{N+1} = z$. Then, by employing the chain rule and the multiplicative property of the determinant, the NLL objective in Eq.(\ref{eq:nll}) can be defined as
\begin{equation}\label{eq:nll1}
\mathcal{L}_{nll}(\boldsymbol{\theta};\mathbf{X},\mathbf{Y})=-\log p_{\mathbf{z}}(\mathbf{z})-\sum_{n=1}^{N}\log\left|\det\dfrac{\partial f_{\boldsymbol{\theta}}^n}{\partial\mathbf{h}^n}(\mathbf{h}^n;\mathbf{X},\mathbf{Y})\right|.
\end{equation}
As a consequence, we only need to ensure that each layer is invertible and corresponding log-determinant of the Jacobian matrix can be efficiently computed, which will be detailed in the following section. Then, the clean HSIs can be sampled from $p_{\mathbf{X}|\mathbf{Y}}(\mathbf{X}|\mathbf{Y},\boldsymbol{\theta_{*}})$ by drawing samples from a simple distribution (e.g. Gaussian) $p_z$ and performing inverse transforms, $\mathop{\mathrm{i.e.,}} \mathbf{X}=f^{-1}_{\boldsymbol{\theta_{*}}}(\mathbf{\hat{z}};\mathbf{Y}), \mathbf{\hat{z}}\sim p_\mathbf{z}$, where $\boldsymbol{\theta_{*}}$ is the learnt parameters of the proposed network.

\subsection{Network Architecture}
\label{sec:network}
In this section, we illustrate the network architecture and implementation details of our proposed method.

\subsubsection{Overall Network Architecture}
% difficulty -> low/high freq -> previous methods -> our encodeer
 While the invertibility of flow-based networks ensures one-to-one mapping, this constraint also imposes limitations on the network design and decreases the fitting ability. Furthermore, the dimensionality of HSIs is significantly larger than RGB images, resulting in the learning of HSI distribution more challenging. Therefore, we propose to decouple the learning of global low-frequency representation and local high-frequency details. Specifically, we propose a flow-based framework namely HIDFlowNet, which is composed of a transformer-based encoder and an invertible decoder as shown in Fig.~\ref{fig:architecture}. The framework employs a conditional encoder without the constraint of invertibility to learn global low-frequency information. Then the flow-based decoder consisting of invertible conditional blocks (ICBs) takes the features maps of the conditional encoder's hidden layers as conditional inputs and transforms samples drawn from Gaussian distribution into local high-frequency information. Since invertible networks are information-lossless and can preserve details \cite{liu2020deep}, the flow-based decoder is ideal for learning the distribution of the high-frequency part of HSIs. Finally, we apply a bilinear upsampling operation to the outputs of the encoder to expand the spatial size. Then the restored HSI is obtained by adding up the outputs of the encoding network and the flow-based decoder so that the global low-frequency and local high-frequency details are restored simultaneously. Next, we will introduce the conditional encoder network and the invertible decoder network in detail.

\subsubsection{Conditional Encoder}
Previous works \cite{dinh2016density, liu2022disentangling, ardizzone2019guided, liu2021invertible} perform either checkerboard pattern squeeze operation or Haar wavelets to reshape image to lower resolutions and capture information in a larger distance when designing invertible networks. However, each time the squeeze operation is performed, the number of channels becomes four times the original number as the size of the image needs to remain unchanged to ensure reversibility. Such operations are not suitable for HSIs which contain tens and even hundreds of spectral bands, as the exponential growth of the number of channels could lead to intolerable computational cost and model complexity. Therefore, inspired by previous work \cite{ma2020decoupling}, we compress the high-dimensional image data by applying down-sampling operations in the encoder which is not necessarily invertible to capture low-frequency information while reducing model complexity in an unsupervised manner. Recently, vision transformers have gained great popularity in various tasks such as classification \cite{chen2021crossvit, he2021spatial, bhojanapalli2021understanding}, segmentation \cite{valanarasu2021medical, chen2021transunet} and image restoration \cite{liang2021swinir, zamir2022restormer}. The self-attention mechanism in transformers enables networks to capture global dependencies and has demonstrated powerful representation capabilities. Therefore, in this work, the encoding network is built by staking a succession of transformers with down-sampling operations to obtain global low-resolution representations as shown in Fig.~\ref{fig:architecture}. Specifically, the locally-enhanced window (LeWin) transformer block proposed in \cite{wang2022uformer} is employed in the HIDFlowNet as the block is considerably efficient and captures both local and global features. Since the LeWin transformer is not the main point of our proposed method, readers could refer to \cite{wang2022uformer} for further details. The downsampling is implemented by a 2-D convolution block with stride=2.

\begin{figure}[h]
  \centering
  \includegraphics[width=\linewidth]{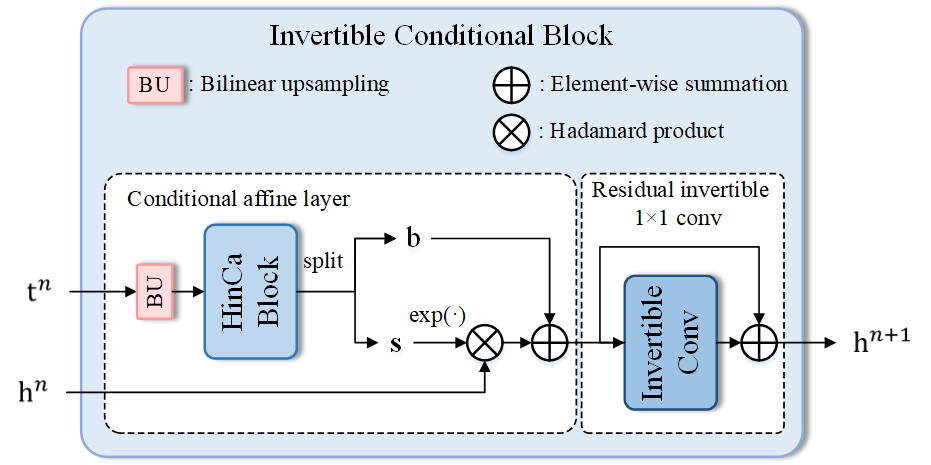}
  \caption{The invertible conditional block is composed of an invertible conditional affine layer and a residual invertible convolution layer. The feature map of the encoder $\textbf{t}^n$ is processed through an upsampling layer and a HinCa Block to generate the scale and bias terms of the affine transform. And then the output $\textbf{h}^{n+1}$ is generated by performing an invertible convolution.}
  \label{fig:ICB}
\end{figure}

\begin{figure}[h]
  \centering
  \includegraphics[width=\linewidth]{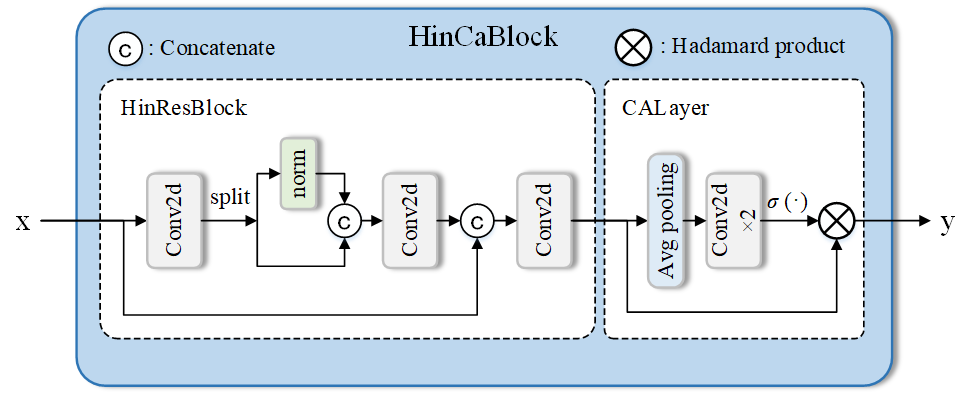}
  \caption{The details of HinCaBlock which consists of a half instance normalization block and a channel attention layer.}
  \label{fig:HinCaBlock}
\end{figure}

\subsubsection{Invertible Decoder}
% As illustrated in Sec. \ref{sec:flow}, the flow-based decoder is composed a sequence of invertible conditional blocks (ICBs) for simpler calculation.
The architecture of the invertible decoder which learns the distribution of high-frequency information requires careful design to ensure that the network is invertible and the Jacobian determinant term in Eq.(\ref{eq:nll}) is tractable. Based on previous works \cite{kingma2018glow, lugmayr2020srflow}, a novel invertible conditional block (ICB) is proposed in this work. As shown in Fig.~\ref{fig:ICB}, each ICB consists of a conditional affine layer and a residual invertible $1 \times 1$ convolution.

The conditional affine layer utilizes an information transfer layer to perform element-wise scaling and addition. Concretely, the conditional affine layer takes the low-resolution feature map $\textbf{t}^n$ of the encoder layer as conditional inputs and generates scale and bias terms, which can be illustrated as
\begin{equation}\label{eq:affine}
\begin{split}
\textbf{s}, \textbf{b} &= \mbox{split}(g_{\boldsymbol{\theta}}(\mbox{BU}(\textbf{t}^n))),\\
\textbf{h}^{n+1} &= \mbox{exp}(\textbf{s}) \odot \textbf{h}^n + \textbf{b},
\end{split}
\end{equation}
where $g_{\boldsymbol{\theta}}$ denotes the information transfer layer, $\mbox{BU}$ denotes bilinear upsampling and $\odot$ is Hadamard product. Half instance normalization block \cite{chen2021hinet} with channel attention \cite{hu2018squeeze} (HinCaBlock) is employed as the information transfer layer in our work, which is shown in Fig.~\ref{fig:HinCaBlock}. Existing works \cite{chen2021hinet,hu2018squeeze} have verified that the HinCaBlock module has strong fitting ability, and we utilize the module to generate the scaling coefficient and the bias of the linear affine transformation. As shown in Fig.~\ref{fig:HinCaBlock}, the HinCaBlock module is composed of two parts: HinResBlock and CALayer. HinResBlock utilizes 3 $\times$ 3 convolutions to generate intermediate features and performs instance normalization on half of the channels to preserve contextual information, which enables the model to ensure the independence between image samples and extract expressive low-level features \cite{chen2021hinet}. CALayer uses channel attention to adaptively enhance the important information in different channels and improves the capability of feature representation of the model.

The Jacobian matrix of this affine transformation is diagonal and the log-determinant can be efficiently computed by adding up the elements of scale $\textbf{s}$. The inverse of this transformation is given by
\begin{equation}\label{eq:inverseaffine}
\textbf{h}^{n} = (\textbf{h}^{n+1} - \textbf{b}) \oslash \mbox{exp}(\textbf{s}),
\end{equation}
where $\oslash$ is element-wise division. \cite{kingma2018glow} proposed an invertible $1 \times 1$ convolution as a permutation operation. However, the determinant of the convolution weight matrix is likely to be a large value and change drastically during the training process as the magnitude of the matrix elements is equivalent. In our work, we further propose a residual invertible $1 \times 1$ convolution to improve the stability of the training process. Specifically, the residual convolution can be defined as
\begin{equation}\label{eq:rc}
\textbf{h}_{ij}^{n+1}=\textbf{W}\textbf{h}_{ij}^n+\textbf{h}_{ij}^n=(\textbf{W}+\textbf{I})\textbf{h}_{ij}^n,
\end{equation}
where $\textbf{h}_{ij}^n$ is the feature vector on spatial coordinate $(i, j)$. The log-determinant is computed in a straightforward way as
\begin{equation}\label{eq:ic determinant}
\log\left|\det\left(\dfrac{d\operatorname{ResidualConv}(\mathbf{h};\mathbf{W})}{d\mathbf{h}}\right)\right|=h\cdot w\cdot\log|\det(\mathbf{W}+\mathbf{I})|,
\end{equation}
where $h$ and $w$ are the height and width of the feature map $\mathbf{h}$, and $\operatorname{ResidualConv}$ is the residual invertible convolution. Since the channel number remains unchanged in the invertible decoder, the log-determinant can be trivially calculated. In addition, the Jacobian determinant term in Eq.(\ref{eq:nll}) prevents the coefficient matrix $\textbf{W}+\textbf{I}$ from being singular. We initialize the parameters $\textbf{W}$ with small values, such that the residual convolution performs as an identity function approximately, which is helpful for training deep networks \cite{kingma2018glow}.

\subsubsection{Objective Function}
As mentioned earlier, we propose a negative log-likelihood loss $\mathcal{L}_{nll}(\boldsymbol{\theta};\mathbf{X},\mathbf{Y})$ to learn the distribution of HSIs. To restore high-quality HSI and accelerate training, we further define reconstruction loss as
\begin{equation}\label{eq:rec}
\mathcal{L}_{rec}(\boldsymbol{\theta};\mathbf{X},\mathbf{Y}, \mathbf{\hat{z}}) = ||f^{-1}_{\boldsymbol{\theta}}(\mathbf{\hat{z}};\mathbf{Y}) - \mathbf{X}||_1,
\end{equation}
where $\mathbf{\hat{z}}$ is a random sample drawn from Gaussian distribution. Finally, the total objective function is defined as
\begin{equation}\label{eq:loss}
\mathcal{L}_{total}(\boldsymbol{\theta};\mathbf{X},\mathbf{Y},\mathbf{\hat{z}}) = \lambda_1 \mathcal{L}_{nll}(\boldsymbol{\theta};\mathbf{X},\mathbf{Y}) + \lambda_2 \mathcal{L}_{rec}(\boldsymbol{\theta};\mathbf{X},\mathbf{\hat{z}})
\end{equation},
where $\lambda_1$ and $\lambda_2$ are hyperparameters. In our experiments, $\lambda_1$ and $\lambda_2$ are set as 0.001 and 1, respectively.

\section{Experiments}
\subsection{Experimental Settings}
In this section, we provide a detailed description of the datasets and training settings in our experiment.
% 31 wavelengths from 450 nm to 700 nm are adopted for training and testing. The code corresponding to the proposed approach is released for reproducibility and future research. \footnote{\url{https://github.com/LiPang/HIDFlowNet}}

\subsubsection{Synthetic Datasets}
Two datasets, i.e., CAVE \cite{park2007multispectral} and KAIST \cite{choi2017high}, are used in our experiments. CAVE dataset consists of 32 HSIs with a spatial resolution of 512 $\times$ 512 over 31 spectral bands. KAIST dataset contains 30 HSIs with a spatial resolution of 2704 $\times$ 3376 over 31 spectral bands. For the CAVE dataset, we use 20 images for training, 2 images for validation and 10 images for testing. For the KAIST dataset, 20 images are used for training and the rest are used for testing. Additionally, 2 images selected from the CAVE dataset are used for validation. We crop the training set with a spatial size of $64 \times 64$ and stride 16 to enlarge training sets, resulting in 16824 training patches in total. Various transformations, i.e., random flipping and multi-angle image rotation (angles of $0^{\circ}$, $90^{\circ}$, $180^{\circ}$, $270^{\circ}$) are used for data augmentation.

% \begin{figure*}[h]
%   \centering
%   \includegraphics[width=0.95\textwidth]{Figures/CAVE.PNG}
%   \caption{Visual result comparison of simulated noise removal on two HSIs selected from CAVE dataset.}\label{fig:cave}
% \end{figure*}

\subsubsection{Real HSI Data} We evaluate all the competing approaches on two real-world noisy HSIs, i.e., Urban dataset\footnote{\url{http://www.tec.army.mil/hypercube}} whose size is $307 \times 307 \times 210$ and Indian Pines dataset\footnote{\url{https://engineering.purdue.edu/∼biehl/MultiSpec/hyperspectral.html}} whose size is $145 \times 145 \times 220$. For computational convenience, the left area of the Urban dataset with a spatial size of $256 \times 256$ and the centre area of the Indian Pines dataset with a spatial size of $128 \times 128$ are cropped for comparison.
% The Indian Pines dataset consists of 145 $\times$ 145 pixels with 220 bands and the Urban dataset consists of 307 $\times$ 307 pixels with 210 bands.

\subsubsection{Noise Setting}
We consider two types of noises (i.e., Gaussian noise and complex noise) which are widely used to simulate real noise situations \cite{zhang2013hyperspectral, chen2017denoising}. In the Gaussian noise case, HSIs are contaminated by noises with variance set as $\{30, 50, 70, 90\}$. In the complex noise case, HSIs are contaminated by non-i.i.d. Gaussian noise, impulse noise, deadlines, strips and mixture noise. Specifically, in the mixture noise case, each band of the clean HSIs is firstly corrupted by Gaussian noise with random intensities which range from 10 to 70. Next, the spectral bands are randomly divided into three parts, each part is added with impulse noise, stripe noise and deadline noise, respectively.

\subsubsection{Competing Methods and Evaluation Metrics}
Nine HSI reconstruction methods are adopted for comparison, including five model-based methods, i.e., BM4D \cite{maggioni2012nonlocal}, LRTDTV \cite{wang2017hyperspectral}, NMoG \cite{chen2017denoising}, FastHyDe \cite{zhuang2018fast}, LLRGTV \cite{he2018hyperspectral}, and four deep learning-based methods, i.e., HSIDCNN \cite{yuan2018hyperspectral}, QRNN3D \cite{wei20203}, SQAD~\cite{pan2022sqad}, SST \cite{li2022spatial}. Three image quality evaluation metrics, including peak signal-to-noise ratio (PSNR), structural similarity (SSIM) \cite{wang2004image} and spectral angle mapper (SAM) \cite{yuhas1993determination}, are employed. Larger values of PSNR and SSIM and smaller values of SAM indicate better image quality.

\subsubsection{Implementation Details}
We implement the proposed framework HIDFlowNet in Pytorch. Adam \cite{kingma2014adam} optimizer with $\beta_1 = 0.9$ and $\beta_2 = 0.999$ is adopted to update model parameters and the learning rate is set to $2 \times 10^{-4}$. All models are trained in an easy-to-difficult way which has been proven helpful for network training \cite{wei20203}. Concretely, the networks are trained with Gaussian noise with random intensities ranging from 10 to 70 for 50 epochs and then trained with mixture noise for another 50 epochs. The training batch size is set as 8. For fair comparisons, all deep learning-based methods are trained and tested in the same way. The models trained for 50 and 100 epochs are employed to remove Gaussian noise and mixture noise respectively. All deep learning-based models are trained on an NVIDIA Geforce RTX 3090 GPU, and it takes approximately 30 hours to complete the training process of our proposed method.

\begin{table*}[t]
\caption{The quantitative denoising results on the CAVE dataset in Gaussian and complex noise cases.}
\label{tab:cave}
\resizebox{\textwidth}{!}{
\begin{tabular}{c|c|c|ccccc|ccccc}
\hline
\multicolumn{1}{l|}{}    & \multicolumn{1}{l|}{} & \multicolumn{1}{l|}{} & \multicolumn{5}{c|}{Model-based methods} & \multicolumn{4}{c}{Deep Learning-based methods} \\ \hline
$\sigma$                 & Index                 & Noisy                 & BM4D & LRTDTV & NMoG & FastHyDe & LLRGTV & HSIDCNN & QRNN3D & SQAD & SST & Ours \\ \hline
\multirow{3}{*}{30}      & PSNR                 & 18.589                & 38.646 & 33.815 & 29.519 & 35.507 & 35.046 & \textbf{39.874} & 38.473 & 37.321 & 39.163 & 38.060 \\
                          & SSIM                  & 0.136                 & 0.937 & 0.876 & 0.658 & 0.914 & 0.892 & 0.961 & 0.944 & 0.923 & \textbf{0.966} & 0.963 \\
                          & SAM                  & 0.994                 & 0.145 & 0.193 & 0.333 & 0.152 & 0.206 & 0.127 & 0.172 & 0.202 & 0.119 & \textbf{0.113} \\ \hline
\multirow{3}{*}{50}      & PSNR                  & 14.152                & 35.790 & 33.002 & 26.795 & 34.464 & 32.532 & \textbf{37.676} & 36.277 & 35.523 & 37.409 & 37.034 \\
                         & SSIM                  & 0.068                 & 0.891 & 0.860 & 0.534 & 0.896 & 0.819 & 0.943 & 0.909 & 0.895 & \textbf{0.952} & 0.951 \\
                         & SAM                   & 1.137                 & 0.192 & 0.209 & 0.415 & 0.172 & 0.274 & 0.158 & 0.227 & 0.225 & 0.139 & \textbf{0.124} \\ \hline
\multirow{3}{*}{70}      & PSNR                  & 11.229                & 33.930 & 32.353 & 24.992 & 33.841 & 30.750 & 36.053 & 33.323 & 33.189 & 36.092 & \textbf{36.130} \\
                         & SSIM                  & 0.041                 & 0.846 & 0.842 & 0.455 & 0.879 & 0.755 & 0.923 & 0.817 & 0.820 & 0.938 & \textbf{0.940} \\
                         & SAM                   & 1.222                 & 0.232 & 0.226 & 0.480 & 0.191 & 0.332 & 0.188 & 0.330 & 0.299 & 0.158 & \textbf{0.135} \\ \hline
\multirow{3}{*}{90}      & PSNR                  & 9.047                & 32.554 & 31.675 & 23.700 & 32.372 & 29.358 & 34.629 & 28.985 & 29.844 & 34.954 & \textbf{35.236} \\
                         & SSIM                  & 0.027                 & 0.806 & 0.826 & 0.403 & 0.846 & 0.700 & 0.897 & 0.600 & 0.647 & 0.922 & \textbf{0.927} \\
                         & SAM                   & 1.279                 & 0.264 & 0.244 & 0.534 & 0.224 & 0.383 & 0.230 & 0.477 & 0.410 & 0.177 & \textbf{0.147} \\ \hline
\multirow{3}{*}{Mixture}  & PSNR  & 13.948 & 18.229 & 32.256 & 19.340 & 18.217   & 24.800 & 34.284          & \textbf{35.341} & 35.003 & 34.484 & 33.535         \\
                          & SSIM  & 0.114  & 0.234  & 0.865  & 0.309  & 0.206    & 0.617  & 0.852           & 0.876          & 0.869  & 0.895  & \textbf{0.899} \\
                          & SAM   & 1.086  & 0.376  & 0.202  & 0.421  & 0.342    & 0.324  & 0.414           & 0.271          & 0.261  & 0.245  & \textbf{0.204} \\ \hline  
\end{tabular}
}
\end{table*}

\begin{table*}[!h]
\caption{Quantitative comparison of denoising performance on the KAIST dataset in Gaussian and complex noise cases.}
\label{tab:kaist}
\resizebox{\textwidth}{!}{
\begin{tabular}{c|c|c|ccccc|ccccc}
\hline
\multicolumn{1}{l|}{}    & \multicolumn{1}{l|}{} & \multicolumn{1}{l|}{} & \multicolumn{5}{c|}{Model-based methods} & \multicolumn{4}{c}{Deep Learning-based methods} \\
\hline
$\sigma$                 & Index                 & Noisy                 & BM4D & LRTDTV & NMoG & FastHyDe & LLRGTV & HSIDCNN & QRNN3D & SQAD & SST & Ours \\
\hline
\multirow{3}{*}{30}      & PSNR                  & 18.589                & 38.672 & 34.299 & 29.244 & 36.829 & 35.188 & \textbf{40.690} & 39.376 & 38.325 & 40.070 & 39.126 \\
                          & SSIM                  & 0.123                 & 0.937 & 0.893 & 0.675 & 0.912 & 0.924 & 0.959 & 0.943 & 0.922 & \textbf{0.963} & 0.951 \\
                          & SAM                   & 0.936                 & 0.142 & 0.177 & 0.328 & 0.155 & 0.172 & \textbf{0.084} & 0.117 & 0.175 & 0.086 & 0.097 \\
\hline
\multirow{3}{*}{50}      & PSNR                  & 14.151                & 35.775 & 32.999 & 26.422 & 34.312 & 32.361 & \textbf{38.622} & 37.282 & 36.585 & 38.529 & 38.174 \\
                          & SSIM                  & 0.060                 & 0.893 & 0.875 & 0.550 & 0.870 & 0.866 & 0.942 & 0.914 & 0.898 & \textbf{0.949} & 0.941 \\
                          & SAM                   & 1.094                 & 0.192 & 0.194 & 0.409 & 0.192 & 0.234 & 0.109 & 0.148 & 0.158 & \textbf{0.103} & 0.104 \\
\hline
\multirow{3}{*}{70}      & PSNR                  & 11.228                & 33.854 & 32.021 & 24.849 & 32.772 & 30.498 & 36.976 & 34.059 & 34.259 & 37.309 & \textbf{37.333} \\
                          & SSIM                  & 0.036                 & 0.850 & 0.856 & 0.474 & 0.823 & 0.808 & 0.921 & 0.822 & 0.828 & \textbf{0.935} & 0.932 \\
                          & SAM                   & 1.186                 & 0.232 & 0.211 & 0.475 & 0.221 & 0.288 & 0.134 & 0.228 & 0.214 & 0.117 & \textbf{0.111} \\
\hline
\multirow{3}{*}{90}      & PSNR                  & 9.047                 & 32.373 & 31.227 & 23.674 & 32.193 & 29.006 & 35.358 & 29.030 & 31.255 & 36.211 & \textbf{36.525} \\
                          & SSIM                  & 0.024                 & 0.810 & 0.838 & 0.425 & 0.809 & 0.758 & 0.888 & 0.572 & 0.697 & 0.919 & \textbf{0.922} \\
                          & SAM                   & 1.249                 & 0.266 & 0.226 & 0.528 & 0.235 & 0.335 & 0.171 & 0.397 & 0.309 & 0.133 & \textbf{0.118} \\
\hline
\multirow{3}{*}{Mixture} & PSNR                  & 13.748                & 17.856 & 32.178 & 18.192 & 17.877 & 24.980 & 34.994 & 36.210 & \textbf{36.617} & 35.764 & 35.057 \\
                          & SSIM                  & 0.103                 & 0.189 & 0.882 & 0.221 & 0.161 & 0.604 & 0.855 & 0.871 & 0.897 & 0.885 & \textbf{0.907} \\
                          & SAM                   & 1.089                 & 0.382 & 0.192 & 0.403 & 0.350 & 0.305 & 0.307 & 0.209 & 0.173 & 0.205 & \textbf{0.139} \\
\hline
\end{tabular}
}
\end{table*}

\begin{figure*}[!h]
  \centering
  \includegraphics[width=0.95\textwidth]{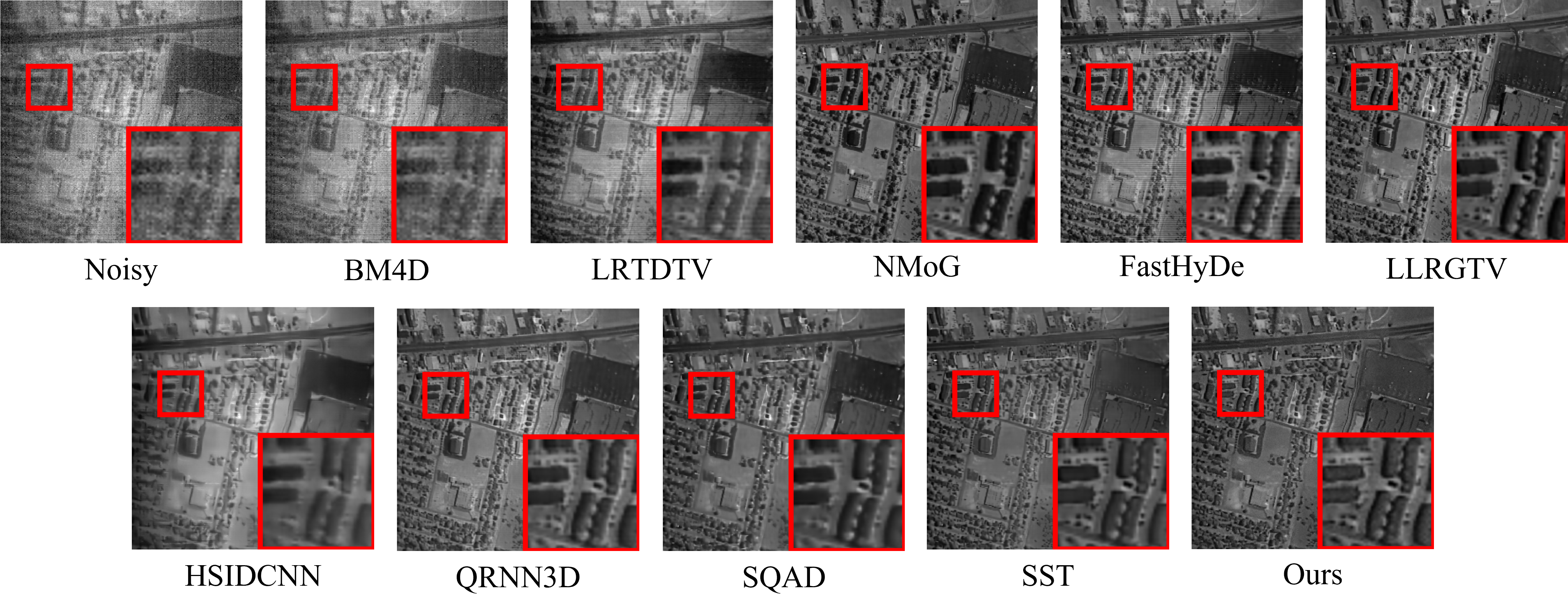}
  \caption{Visual comparison of denoising results on the 104th band in Urban dataset.}\label{fig:realworld_urban}
\end{figure*}

\subsection{Experimental Results}
\subsubsection{Experiment on Synthetic Data}
The denoising results on the CAVE dataset are shown in Table \ref{tab:cave}. It can be seen that our proposed HIDFlowNet can obtain better performance in most cases. While achieving desirable results in Gaussian noise cases, most model-based methods fail to tackle complex noise as manually designed priors cannot fully describe complex situations. In addition, although HSIDCNN performs best with respect to PSNR in several cases by performing multiscale feature extraction, our HIDFlowNet also achieves promising PSNR value and performs significantly better with respect to other evaluate indexes. Model-based approaches yield either noisy images or over-smooth results. Deep learning-based methods obtain promising denoising results but are also prone to provide over-smooth predictions since these methods overemphasize the pixel similarity and ignore the underlying distribution of clean HSIs. In contrast, HIDFlowNet is more capable of preserving fine-grained details while restoring spatial smoothness without introducing undesirable artifacts. The excellent performance of HIDFlowNet is primarily owing to the fact that the compressive encoding component suppresses noise and enhances the low-frequency part of HSIs, and the flow-based decoder enjoys the information-less property and preserves textural details. Moreover, HIDFlowNet also exhibits desirable denoising performance on the KAIST dataset as shown in Table \ref{tab:kaist}, which further verifies the superiority of our method.
% The visualization results of reconstructed HSIs are shown in Figure \ref{fig:cave}. As can be seen, model-based approaches
% In addition, the spectral density curves are also provided to demonstrate the spectral consistency between the restoration results of HIDFlowNet and ground truth.
% In addition, PSNR values across the spectrum are also provided to demonstrate that our proposed method outperforms almost all bands (see Figure\ref{fig:teaser}).

% \begin{figure*}[!h]
%   \centering
%   \includegraphics[width=1\textwidth]{Figures/urban_curves.png}
%   \caption{Visual comparison of the spectral signatures of point (185,132) in Urban dataset. }\label{fig:realworld_urban_curves}
% \end{figure*}

\begin{figure*}[!h]
  \centering
  \includegraphics[width=0.95\textwidth]{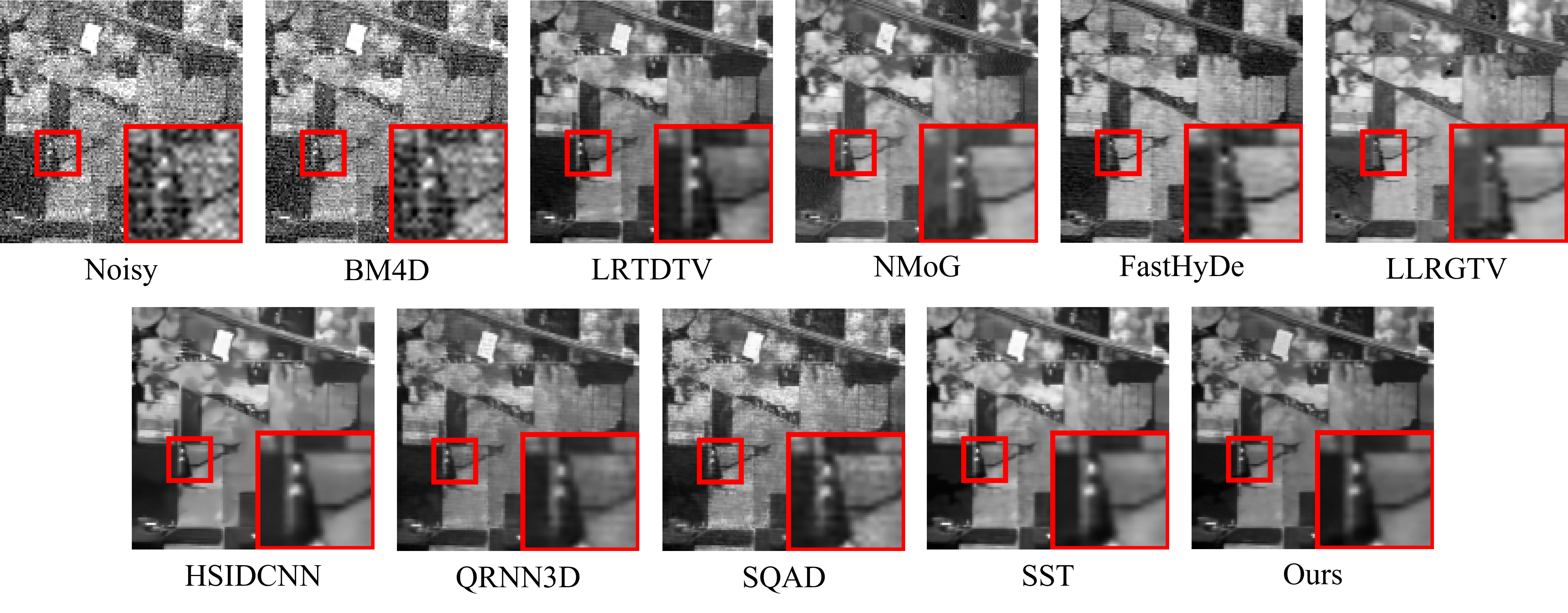}
  \caption{Visual comparison of denoising results on the 5th band in Indian Pines dataset.}\label{fig:realworld_indian}
\end{figure*}

\subsubsection{Experiment on Real-World Data}
This section illustrates the comparison results of all the methods on the real HSI data. Specifically, all the deep learning-based methods trained on the KAIST dataset are tested on the Urban and Indian Pine datasets. Since there is no ground truth for real data, we provide visualization results for comparison. Fig.~\ref{fig:realworld_urban} shows the denoising results of the 104th band for the Urban dataset. It can be observed that the original image is seriously degraded owing to environmental factors such as terrible atmosphere or sensor failure. The denoised results of BM4D, LRTDTV and FastHyDe contain obvious noise residue as there is a serious discrepancy between manually designed priors such as i.i.d. Gaussian noise assumption and the real situation. While NMoG and LLRGTV achieve desirable denoising performance, these methods still suffer oversmooth issues. Additionally, the denoising results of the deep learning-based approaches, e.g., HSIDCNN, QRNN3D and SQAD, also contain some oversmooth areas. As for our proposed HIDFlowNet, it performs best both on noise removal and texture preservation. Similar denoising results on Indian Pines dataset are provided in  Fig.~\ref{fig:realworld_indian}.

\subsubsection{Effectiveness of Flow Model}
\begin{figure}[h]
  \centering
  \includegraphics[width=\linewidth]{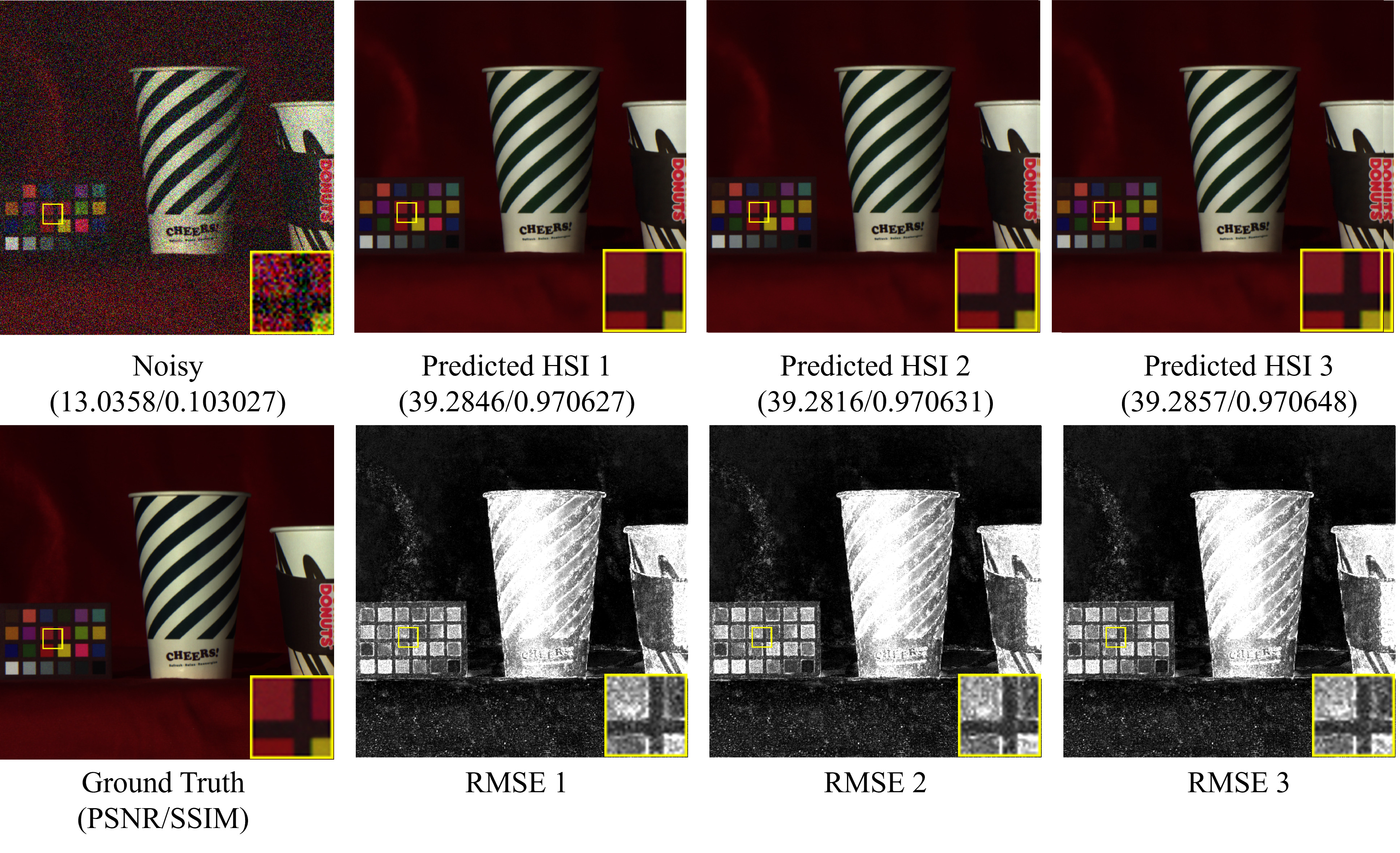}
  \caption{Diverse predictions of clean HSI given one noisy HSI in the KAIST dataset by our method.}\label{fig:flow}
\end{figure}

We present visualization results of the generated HSIs derived from different Gaussian noises in Fig.~\ref{fig:flow} to verify the effectiveness of our proposed flow-based model. It can be observed that while generated HSIs are highly similar which verifies the stability of the trained model, there still exist differences in local details owing to different noises, confirming the effectiveness of our proposed flow-based model. In addition, to further verify the stability of our model, given a noisy image from the CAVE dataset as a condition, we obtain 20 different clean images by sampling 20 times from the standard normal distribution and conducting inverse transformations. For the 20 reconstructed images, we calculated the mean, standard deviation, and range (i.e., the difference between the maximum and minimum values) of PSNR and SSIM values. The results are shown in Table \ref{tab:stability}. It can be seen that the PSNR values of the generated images are stable between 41.6 and 41.8, and the SSIM values are stable between 0.97497 and 0.97521. All the generated images are of high quality, proving the effectiveness of our method.

\subsection{Ablation Study}
In this section, we provide an ablation study on the components of HIDFlowNet and model complexity.

\subsubsection{Feature Decoupling Analysis}
\begin{figure}[h]
  \centering
  \includegraphics[width=\linewidth]{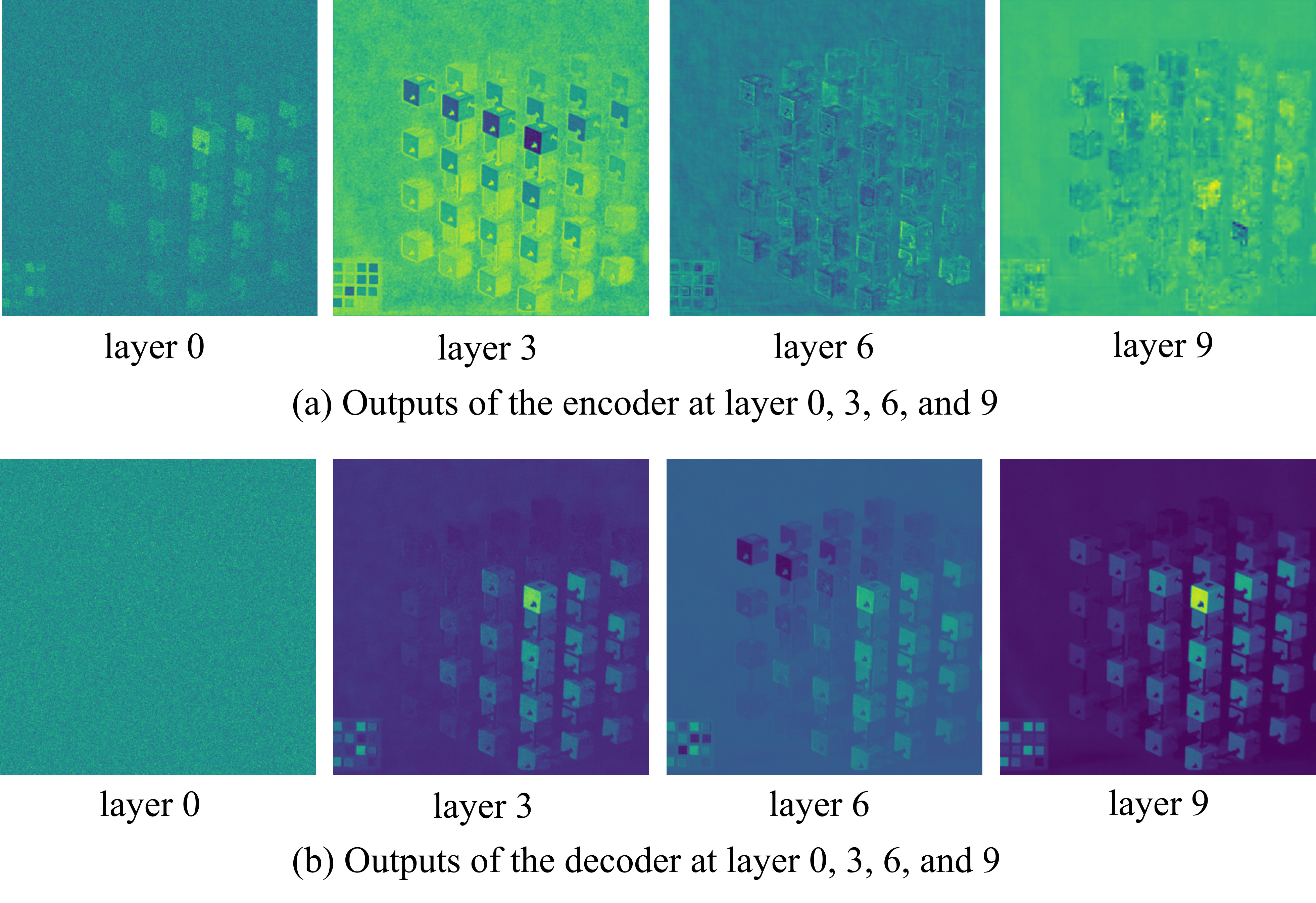}
  \caption{The visual results of the feature maps of the conditional encoder and the invertible decoder. It can be seen that with the increase of layers, the outputs of the encoder tend to ignore local details (e.g., the joint of the blocks) and gradually capture global low-frequency information.}\label{fig:Feature Decoupling}
\end{figure}

In addition to quantitative results, we provide visual analysis to further prove the effectiveness of the proposed encoding network and the flow-based decoder. Specifically, the inputs and the feature maps of the 3th, 6th and 9th layers of the encoder and decoder are depicted in Fig. \ref{fig:Feature Decoupling}. It can be seen that with the increase of layers, the outputs of the encoder tend to ignore local details (e.g., the joint of the blocks) and gradually capture global low-frequency information. Since attention is calculated in local windows as elaborated in \cite{wang2022uformer}, the feature map of the last layer exhibits a relatively obvious reticular structure. The outputs of the decoder demonstrate that with the guidance of the encoder, random Gaussian noise is transformed into local high-frequency information progressively, convincing the feasibility of the invertible network.

% \begin{figure*}[h]
%   \centering
%   \includegraphics[width=0.8\textwidth]{Figures/samples.png}
%   \caption{Diverse predictions of clean HSI given one noisy HSI in the KAIST dataset by our method.}\label{fig:flow}
% \end{figure*}

\begin{table}[h]
\centering
\caption{The image quality statistics of 20 sampled clean images. All the generated images are of high quality, proving the stability of our proposed model.}
\label{tab:stability}
\renewcommand\arraystretch{0.85}
\setlength{\tabcolsep}{1.5mm}{
\begin{tabular}{cccc}
\toprule
     & Mean    & Standard Deviation & Range   \\ \midrule
PNSR & 41.756  & 0.004              & 0.014   \\
SSIM & 0.97500 & 0.00002            & 0.00005  \\
\bottomrule
\end{tabular}
}
\end{table}

\subsubsection{Component Analysis}
There are two components in an invertible conditional block, including an affine conditional layer and a residual invertible convolution. In this section, to verify the effectiveness and rationality of the two components adopted in our work, we conduct denoising on the KAIST dataset in Gaussian noise case with $\sigma=50$ for comparison and the effectiveness of the two components is explored as illustrated in Table \ref{tab:component}. As can be seen, the model without affine conditional layers demonstrates the worst performance since the decoder is a pure generative network without conditional information in this case, and the quality of the denoising result is highly reliant on the performance of the encoder. HIDFlowNet adopted in our work outperforms other configurations, verifying the rationality of the proposed approach. % Furthermore, HIDFlowNet adopted in our work outperforms that without either residual invertible convolutions or residual structure, verifying the rationality of the proposed approach. and the residual structure of the invertible convolution

% \begin{table}[h]
% \caption{Ablation study of the two components in the invertible conditional block.}
% \label{tab:component}
% \renewcommand\arraystretch{0.85}
% \setlength{\tabcolsep}{1mm}{
% \begin{tabular}{lccc}
% \toprule
% Configuration                          & PSNR   & SSIM  & SAM   \\ \midrule
% No Invertible Conditional Affine Layer & 32.145 & 0.896 & 0.150 \\
% No Residual Invertible Convolution     & 37.837 & 0.940 & 0.108 \\
% Ours                                   & \textbf{38.067} & \textbf{0.942} & \textbf{0.101} \\ \bottomrule
% \end{tabular}
% }
% \end{table}

\subsubsection{Objective Function Analysis}
To further investigate the effectiveness of the negative log-likelihood (NLL) loss and reconstruction loss introduced in Sec. \ref{sec:network}, we conduct model training on the KAIST dataset with different objectives and test the models on the KAIST test set in Gaussian noise case with $\sigma=50$. The results are shown in Table \ref{tab:objective} and Table \ref{tab:objective diversity comparison}. From Table \ref{tab:objective} we can see that the model trained with the NLL loss outputs low-quality images since it is difficult to learn the distribution of HSIs owing to the limitation of HSI dataset size, diversity of scene images and high dimensionality of HSIs. There is no significant image quality difference when the model is trained with reconstruction loss or combined loss. However, the model trained with reconstruction loss is equivalent to deterministic models taking the noisy HSI and random Gaussian noise as inputs and outputting a single clean HSI. Therefore, the model does not learn the distribution of the HSIs and is incapable of generating diverse clean HSIs. In contrast, our proposed method learns the conditional distribution of clean HSIs by minimizing the NLL loss and improves image quality by minimizing the reconstruction loss. We further sample 20 clean HSIs given a noisy image from the KAIST dataset as a condition and calculate the statistics of PSNR values as shown in Table \ref{tab:objective diversity comparison}. From the table, it can be seen that the model trained with reconstruction loss outputs high-quality images but fails to generate diverse images, while the model trained with NLL loss exhibits the opposite behaviour. Compared with the other two models, the model trained with combined loss which is adopted in our work is able to generate diverse high-quality images, verifying the rationality of our proposed method.

\begin{table}[h]
\centering
\caption{Ablation study of the two components in the invertible conditional block. RIC indicates residual invertible convolution and ICAL indicates invertible conditional affine layer. Configuration \uppercase\expandafter{\romannumeral1} indicates using RIC, \uppercase\expandafter{\romannumeral2} indicates using ICAL and \uppercase\expandafter{\romannumeral3} indicates using both.}
\label{tab:component}
\renewcommand\arraystretch{1}
\setlength{\tabcolsep}{2mm}{
\begin{tabular}{c|cc|ccc}
\hline
\multicolumn{1}{l|}{Configuration}     & RIC      & ICAL      & PSNR   & SSIM  & SAM   \\ \hline
\uppercase\expandafter{\romannumeral1} & \ding{51} & \ding{55} & 32.145 & 0.896 & 0.150 \\
\uppercase\expandafter{\romannumeral2} & \ding{55} & \ding{51} & 37.837 & 0.940 & 0.108 \\
Ours & \ding{51} & \ding{51} & \textbf{38.174} & \textbf{0.941} & \textbf{0.104} \\ \hline
\end{tabular}
}
\end{table}

% NLL indicates negative log-likelihood loss and Rec indicates reconstruction loss.
\begin{table}[h]
\centering
\caption{Denoising performance comparison of the models trained with different objectives. Configuration \uppercase\expandafter{\romannumeral1} indicates using NLL, \uppercase\expandafter{\romannumeral2} indicates using reconstruction loss and \uppercase\expandafter{\romannumeral3} indicates using both.}
\label{tab:objective}
\renewcommand\arraystretch{1}
\setlength{\tabcolsep}{2mm}{
\begin{tabular}{c|cc|ccc}
\hline
Configuration                          & \multicolumn{1}{c}{NLL}   & \multicolumn{1}{c|}{Rec}   & PSNR            & SSIM           & SAM            \\ \hline
\uppercase\expandafter{\romannumeral1} & \multicolumn{1}{c}{\ding{51}} & \multicolumn{1}{c|}{\ding{55}} & 30.459          & 0.782          & 0.141          \\
\uppercase\expandafter{\romannumeral2} & \ding{55}                         & \ding{51}                          & 37.930          & \textbf{0.953} & \textbf{0.096} \\
Ours & \ding{51}                         & \ding{51}                          & \textbf{38.174} & 0.941          & 0.104          \\ \hline
\end{tabular}
}
\end{table}

\begin{table}[h]
\centering
\caption{The image quality statistics of 20 sampled images generated by different models. Our proposed model trained with combined loss is able to generate diverse high-quality images, verifying the effectiveness of our proposed method. Configuration \uppercase\expandafter{\romannumeral1} indicates using NLL, \uppercase\expandafter{\romannumeral2} indicates using reconstruction loss and \uppercase\expandafter{\romannumeral3} indicates using both.}
\label{tab:objective diversity comparison}
\renewcommand\arraystretch{1}
\setlength{\tabcolsep}{0.45mm}{
\begin{tabular}{c|cc|ccc}
\hline
Configuration                          & NLL       & Rec       & Mean   & Standard Deviation & Range    \\ \hline
\uppercase\expandafter{\romannumeral1} & \ding{51} & \ding{55} & 30.409 & 0.01693           & 0.06060 \\
\uppercase\expandafter{\romannumeral2} & \ding{55} & \ding{51} & 37.606 & 0.00001           & 0.00004 \\
Ours & \ding{51} & \ding{51} & 37.996 & 0.00313           & 0.01281 \\ \hline
\end{tabular}
}
\end{table}

\begin{table}[!h]
\centering
\caption{Ablation study of different network depth.}
\label{tab:depth}
\renewcommand\arraystretch{0.85}
\setlength{\tabcolsep}{1.3mm}{
\begin{tabular}{cccccc}
\toprule
Depth & PSNR            & SSIM           & SAM            & Parameters (M)        & Time (s)\\ \midrule
6     & 37.779          & 0.940          & 0.102          &\textbf{1.937}	      &\textbf{0.374}\\
9     & 38.174          & 0.941          & 0.104     &2.808	              &0.467\\
12    & \textbf{38.315} & \textbf{0.944} & \textbf{0.101} &3.679	              &0.628\\ \bottomrule
\end{tabular}
}
\end{table}

\subsection{Limitation}
While our proposed HIDFlowNet exhibits plausible denoising performance, there are still several limitations. Specifically, the invertible requirement of flow-based models puts limitations on the use of various operations such as convolution with larger kernels, attention mechanisms and dimension reduction, reducing the fitting ability of the network. Moreover, the proposed method lacks control over the generative process and is unable to explicitly generate HSIs with expected specific properties such as higher SSIM. In the future, novel invertible frameworks and controllable generative models are worth further exploration to alleviate these problems.

% In addition, the Jacobian matrix of the framework needs to be tractable, presenting further challenges when designing flow-based models.

\section{Conclusion}
To alleviate the ill-posed nature of HSI denoising (i.e., multiple predictions are reasonable for a given noisy HSI) which is ignored by most existing deep learning-based approaches, this paper proposes a novel flow-based network namely HIDFlowNet. The network directly learns the distribution of clean HSIs conditioned on noisy counterparts and is capable of generating diverse clean HSIs. Specifically, the proposed HIDFlowNet is composed of a conditional encoder and an invertible decoder to decouple the learning of low-frequency and high-frequency information. The encoder utilizes transformers and down-sampling operations to obtain low-resolution images so that global representation is effectively extracted, while the decoder employs a series of invertible conditional blocks to preserve local details. Extensive experiments on two synthetic datasets and one real-world dataset demonstrate the superiority of our proposed model both quantitatively and qualitatively.

\bibliographystyle{ACM-Reference-Format}
\bibliography{sample-base.bib}

\end{document}